\documentclass[runningheads]{llncs}
\usepackage{graphicx}
\usepackage{soul}
\usepackage{xcolor}
\usepackage{enumitem}
\usepackage{url}
\usepackage{hyperref}
\makeatletter
\newcommand{\printfnsymbol}[1]{%
  \textsuperscript{\@fnsymbol{#1}}%
}
\makeatother

\begin{document}
\title{Towards Active Learning Based Smart Assistant for Manufacturing}
\titlerunning{Towards AL-based Smart Assistant for Manufacturing}
%
\author{
    Patrik Zajec\inst{1,2}\thanks{equal contribution} \and 
    Jo\v{z}e M. Ro\v{z}anec\inst{1,2,3}\printfnsymbol{1}\orcidID{0000-0002-3665-639X} \and
    Inna Novalija\inst{2}\orcidID{0000-0003-2598-0116}
    \and Bla\v{z} Fortuna\inst{2,3} \and 
    Dunja Mladeni\'{c}\inst{2} \and
    Klemen Kenda\inst{1,2,3}\orcidID{0000-0002-4918-0650}
}

\authorrunning{P. Zajec et al.}

\institute{Jo\v{z}ef Stefan International Postgraduate School, Jamova 39, 1000 Ljubljana, Slovenia,\\
\email{joze.rozanec@ijs.si},
\and
Jo\v{z}ef Stefan Institute, Jamova 39, 1000 Ljubljana, Slovenia
\and 
Qlector d.o.o., Rov\v{s}nikova 7, 1000 Ljubljana, Slovenia}

\maketitle              
\begin{abstract}
A general approach for building a smart assistant that guides a user from a forecast generated by a machine learning model through a sequence of decision-making steps is presented. We develop a methodology to build such a system. The system is demonstrated on a demand forecasting use case in manufacturing. The methodology can be extended to several use cases in manufacturing. The system provides means for knowledge acquisition, gathering data from users. We envision active learning can be used to get data labels where labeled data is scarce. 

\end{abstract}

\section{Introduction}\label{INTRODUCTION}

The increasing digitalization of manufacturing has accelerated the information flow. Technologies such as Cyber-Physical Systems (CPS), Industrial Internet of Things (IIoT), and Artificial Intelligence (AI) are bringing an extensive added value into Industry 4.0 value chains \cite{grangel:2019}. In particular, AI has been successfully researched and applied on several manufacturing tasks (e.g., predictive maintenance, production simulation, and production planning). The advancement of AI and its applications in manufacturing is conditioned by data availability. Though much data is available from software, such as Enterprise Resource Planning (ERP) or Manufacturing Execution Systems (MES), many aspects are not captured by sensors or such software. An example of such information is the collective knowledge, which employees are aware of, but is not captured by existing software integrations. To mitigate this data gap, we propose a software solution with a user interface to collect locally observed collective knowledge~\cite{bradevsko2017curious}. An example of such data acquisition can be feedback collection regarding forecasts, forecast explanations, or decision-making options. Feedback can be provided in an implicit (e.g., by not editing an option) or explicit form (e.g., marking a forecast explanation to be improbable). Another example can be asking the user to input yet unknown data (e.g., a decision taken in a certain context that was not registered in the past). When recommending decision-making options to the user, many decisions must be made: which subset of available decision-making options to display, how to rank them, or how to enable the user to provide useful feedback. We envision active learning to weigh which decision-making options are most informative to the system when users' feedback is provided. 

We demonstrate a conceptual design and a developed system that can acquire and encapsulate complex knowledge. The system is based on semantic technologies, considering ontology concepts that are generic and ported to multiple use cases. We demonstrate its usability on demand forecasting, providing recommendations for transport scheduling. The system integrates demand forecasting models, explainable AI (XAI), a decision-making recommender system, and a knowledge graph. The aforementioned components are used to develop decision-making workflows, which are displayed through an interactive user interface. Feedback is collected from users regarding forecasts, forecast explanations, and decision-making options displayed to the users.

\section{Related Work}\label{RELATED-WORK}

In smart manufacturing, several characteristics (such as context awareness, modularity, heterogeneity, and interoperability) and technologies (such as intelligent control, energy efficiency, cybersecurity, CPS, IoT, data analytics, and IT-based production management) have been identified to play a crucial role \cite{mittal:2019}. 
Variety and depth of technologies in Industry 4.0 represent a great potential, and micro-level local units' usage provides the best outcomes \cite{buchi:2020}.
Several risks are associated with the implementation of smart technologies in manufacturing \cite{micheler:2019} such as i) the perceived risk of novel technologies, ii) the complexity of integration, and iii) the consideration of human factors. 
As several most accurate AI algorithms (such as gradient boosting or deep neural networks) are difficult to explain and justify, they represent a significant perceived risk of novel technologies in a shop-floor environment.

In the context of manufacturing, explainable AI (XAI) technologies \cite{arrieta2020explainable} have been tested in several scenarios such as predictive maintenance \cite{hrnjica:2020}, real-time process management \cite{rehse2019towards} and quality monitoring \cite{goldman2021explaining}.
One of our research goals is to highlight the explainability of the algorithms and methods used in smart manufacturing processes, aligning XAI technologies with human interaction. We also aim to collect feedback on the quality of such explanations, since there are few validated measurements for user evaluations on explanations' quality \cite{van2021evaluating}.

Active Learning (AL) is usually the natural approach to provide human-in-the-loop functionalities in advanced AI systems.
Typically, AL attempts to improve learners' performance by asking questions to an expert (e.g., query a human operator) to obtain labels for data instances~\cite{settles2009active}. 
Since users are usually reluctant to provide information and feedback, AL is used to identify a set of data instances on which the provided users' input conveys the most valuable information to the system~\cite{elahi2016survey}. 
In a decision-making process in manufacturing, AL can also be implemented in recommender systems. In such cases, it tackles obtaining high-quality data that better represents the user's preferences and improves the recommendation quality. 
The ultimate goal is to acquire additional feedback that enables the system to generate better recommendations~\cite{elahi2016survey}.
Collecting feedback from forecast explanations can be realized with a framework of three components: a forecasting engine, an explanation engine, and a feedback loop to learn from the users~\cite{tulli2020learning}. 
We extend this approach to collect feedback from forecasts, forecast explanations and decision-making options we recommend to the users. 

\section{Proposed Methodology and System}\label{METHODOLOGY}

To realize a system described in Section~\ref{INTRODUCTION}, we developed a methodology to identify relevant components, decision-making options, information and feedback of interest, and how to collect them. The methodology consists of ten steps:
\begin{enumerate}
    \item create an AI model, to provide predictions that comply with a given use case;
    \item provide local forecast explanations, either by querying a glass-box model or using complementary methods for black-box models;
    \item list decision-making options available to the user;
    \item create a flow of decision-making options available to the user;
    \item list the kind of feedback expected for forecasts, forecast explanations, and decision-making options. Identify opportunities for implicit feedback;
    \item create a list of relevant entities related with forecasts, explanations, and decision-making options;
    \item extend the ontology~\cite{DVN/UGYHLP_2021} to include use-case specific entities and relate them to the entities that model the AI model, forecasts, forecast explanations, and feedback;
    \item instantiate a knowledge graph based on the ontology entities;
    \item develop a software application binding forecasts, forecast explanations, and decision-making options while enabling decision-making flows and feedback gathering;
    \item develop an Active Learning module that receives input from the database and knowledge graph to suggest data instances that are expected to be most informative to the system. This input can be used by the decision-making recommender system;
\end{enumerate}

\begin{figure*}[!t]
\centering
\includegraphics[width=4.3in,height=2in]{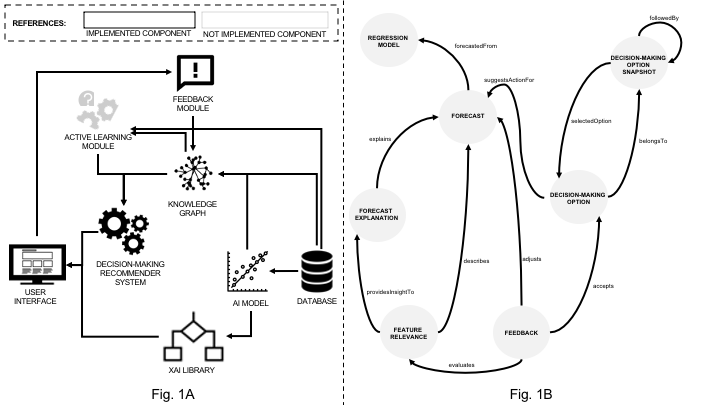}
\caption{Fig.~\ref{F:DIAGRAM}A displays a diagram of the system components and their interaction. Fig.~\ref{F:DIAGRAM}B shows the main ontology concepts we considered, and their relationships.}
\label{F:DIAGRAM}
\end{figure*}

\noindent The system requires at least eight components (see Fig.~\ref{F:DIAGRAM}A): 
\begin{itemize}
    \item \textbf{Database}, stores operational data from the manufacturing plant. Data can be obtained from ERP, MES, or other manufacturing platforms;
    \item \textbf{Knowledge Graph}, stores data ingested from a database or external sources and connects it, providing a semantic meaning. To map data from the database to the knowledge graph, virtual mapping procedures can be used, built considering ontology concepts and their relationships;
    \item \textbf{Active Learning module}, aims to select data instances whose labels are expected to be most informative to the system and thus help enhance AI model's (e.g., predictive model or recommender system model) performance. Obtained labels are persisted to the knowledge graph and database;
    \item \textbf{AI model}, aims to solve a specific task relevant to the use case, such as classification, regression, clustering, or ranking;
    \item \textbf{XAI Library}, provides some insight into the AI model's rationale used to produce the output for the input instance considered at the task at hand. E.g., in the case of a classification task, it may indicate the most relevant features for a given forecast;
    \item \textbf{Decision-Making Recommender System}, recommends decision-making options to the users. Recommended decision-making options can vary depending on the users' profile, specific use case context, and feedback provided in the past;
    \item \textbf{Feedback module}, collects feedback from the users and persists it into the knowledge graph;
    \item \textbf{User Interface}, provides relevant information to the user through a relevant information medium. The interface must enable user interactions to create two-way communication between the human and the system.
\end{itemize}

The knowledge graph is a central component of the system. Instantiated from an ontology (see Fig.~\ref{F:DIAGRAM}B), it relates forecasts, forecast explanations, decision-making options, and feedback provided by the users. To ensure context regarding decision-making options and feedback provided is preserved, different relationships are established. The feedback entity directly relates to a forecast, forecast explanation, and decision-making option. While a chain of decisions can exist for a given forecast, there is a need to model the decision-making options available at each stage and the sequence on which they are displayed. To that end, the decision-making snapshot entity aims to capture a list of decision-making options provided at a given point in time. A relationship between decision-making option snapshots (\textit{followedBy}) provides information on such a sequence. For each decision-making snapshot, a \textit{selectedOption} relationship is created to the user's selected decision-making option. To link the first decision-making options to the forecast, the \textit{suggestsActionFor} relationship is created between the forecast entity and entities that correspond to the first decision-making options displayed for that forecast. Since the decision-making options are linked to decision-making option snapshot and preserve a sequential relationship, all decision-making options can be traced back to the forecast that originated them.

\section{Use Case}\label{USE-CASE}

\begin{figure*}[!t]
\centering
\includegraphics[width=4.3in]{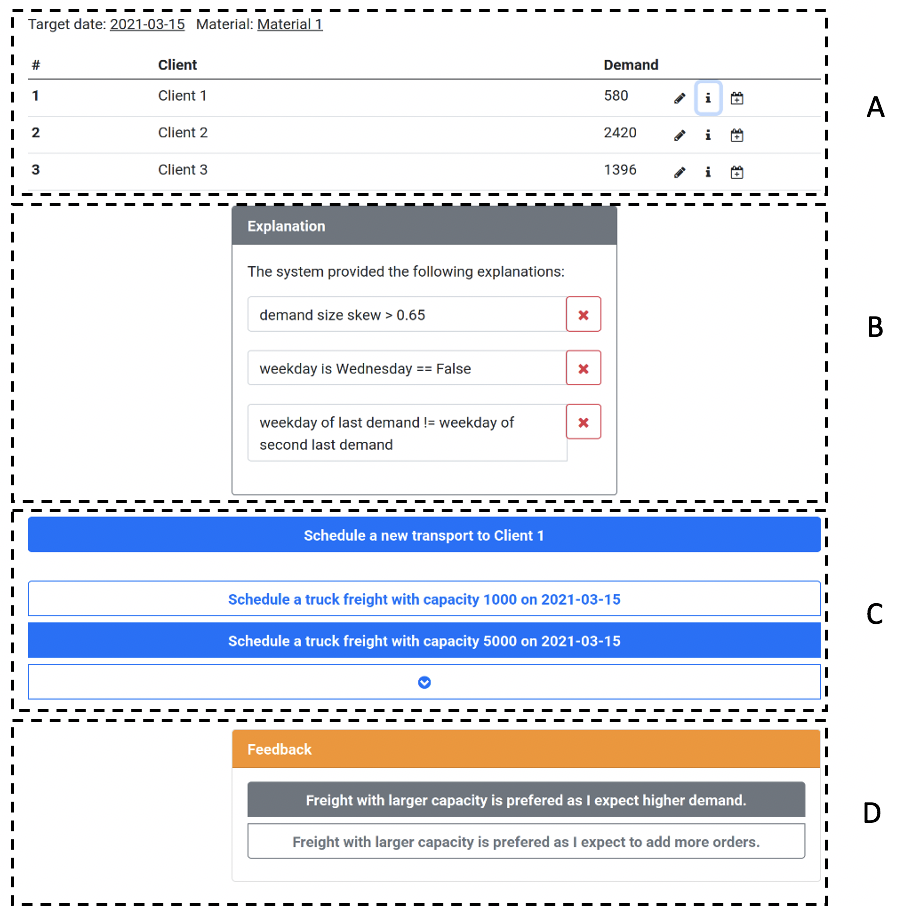}
\caption{User interface, displaying forecasts, forecast explanations, and recommended decision-making options.}
\label{F:USER-INTERFACE}
\end{figure*}

Demand forecasting is a key component of supply chain management since it directly affects production planning and order fulfillment. Accurate forecasts enable operational and strategic decisions regarding manufacturing and logistics for deliveries. For the evaluation of our methodology, we developed a model to forecast demand on a material and client level daily. The model was trained on three years of data for 516 time-series corresponding to 279 materials and 149 clients of a European automotive original equipment manufacturer's daily demand. To validate the system we developed, a subset of demand forecasts was used. Forecast explanations were generated with the LIME library ~\cite{ribeiro2016should}. The initial decision-making options recommender system was implemented with a set of simple heuristics that allow to select a new transport or chose among existing ones that satisfy certain criteria (e.g., have enough capacity to satisfy the expected demand for a given client). Finally, a user interface was developed to display forecasts, forecast explanations, and decision-making options (see Fig.~\ref{F:USER-INTERFACE}). In the user interface, we identify four distinct parts:
\begin{enumerate}[label=\Alph*]
    \item \textbf{Forecast panel}: given date and material, it displays the forecasted demand for different clients. For each forecast, three options are available: edit the forecast (providing explicit feedback on the forecast value), display the forecast explanation, and display the decision-making options. The lack of editing on displayed forecasts is considered implicit feedback approving the forecasted demand quantities.
    \item \textbf{Forecast explanation panel}: displays the forecast explanation for a given forecast. Our implementation displays the top three features identified by the LIME algorithm as relevant to the selected forecast. If the user considers some of the features displayed do not explain the given forecast, they can provide feedback by removing it from the list.
    \item \textbf{Decision-making options panel}: displays possible decision-making options for a given forecast or step in the decision-making process.
    \item \textbf{Feedback panel}: gathers feedback from the user to understand the reasons behind the chosen decision-making option. While some pre-defined are shown to the user, we always include the user's possibility to add their own reasons and enrich the existing knowledge base. Such data can be used to expand feedback options displayed to the users in the future.
\end{enumerate}

Though the current decision-making options recommender system is constrained to heuristics, we envision that in the future, more complex models can be developed leveraging data regarding user interactions with the recommendations we currently display.

\section{Evaluation and Discussion}\label{EVALUATION}

To evaluate the proposed methodology and system, empirical evaluations and the development of a concrete use case were utilized \cite{grangel:2019}. In particular, we applied the methodology outlined in Section~\ref{METHODOLOGY} (except for the Active Learning module implementation) to develop a system that enables displaying forecasts, forecast explanations, and decision-making options, provide a decision-making workflow, and means to collect feedback and knowledge from the users\footnote{A video demonstrating the application is available at \url{https://youtu.be/Kx5UnE_yTM0}}. The system we developed for the particular use case of demand forecasting proves that a semantic approach enables effective and flexible means to solve complex knowledge acquisition and solve interoperability conflicts. Since the system was not deployed to production environments, we cannot assess the perceived quality of the forecast explanations, and the impact of the given forecast explanations and decision-making options provided. The impact of the forecasting models was assessed on data provided by EU H2020 FACTLOG and STAR project partners~\cite{rozanecACM}. The development of an Active Learning module remains the subject of future work.

\section{Conclusion and Future Work}
The current work presents a system's conceptual design to acquire and encapsulate complex knowledge using semantic technologies and AI. The system was instantiated for the demand forecasting use case in the manufacturing domain, using real-world data from partners from the EU H2020 projects STAR and FACTLOG. In particular, the system provides forecasts, forecast explanations, decision-making options, and the capability to provide implicit and explicit feedback. The system enables the development of an active learning module that can enhance data collection by identifying promising data instances that, when labeled, are expected to be most informative to the system.
Future work will focus on implementing an active learning module and explore recommender systems that learn from data to provide decision-making options to the users.

\section*{Acknowledgements}
This work was supported by the Slovenian Research Agency and the European Union’s Horizon 2020 program projects FACTLOG under grant agreement H2020-869951 and STAR under grant agreement number H2020-956573.

\bibliographystyle{splncs04}
\bibliography{bibliography}
\end{document}